\documentclass{article}

\PassOptionsToPackage{numbers, compress}{natbib}

\usepackage[preprint]{neurips_2024}

\usepackage[utf8]{inputenc} 
\usepackage[T1]{fontenc}    
\usepackage{hyperref}       
\usepackage{url}            
\usepackage{booktabs}       
\usepackage{amsfonts}       
\usepackage{nicefrac}       
\usepackage{microtype}      
\usepackage{xcolor}         
\usepackage{amsmath}
\usepackage{amssymb}
\usepackage{mathtools}
\usepackage{amsthm}
\usepackage{todonotes}

\usepackage{cleveref}
\usepackage{algorithm}
\usepackage{algorithmic}
\usepackage{natbib}

\usepackage{listings}
\newcommand\nnfootnote[1]{%
  \begin{NoHyper}
  \renewcommand\thefootnote{}\footnote{#1}%
  \addtocounter{footnote}{-1}%
  \end{NoHyper}
}

\title{STMA: A Spatio-Temporal Memory Agent for Long-Horizon Embodied Task Planning}

\author{%
Mingcong Lei$^{135*}$, Yiming Zhao$^{415*}$,  Ge Wang$^{15}$, Zhixin Mai$^{135}$ \\ \textbf{Shuguang Cui$^{213}$, Yatong Han$^{135\dagger}$, Jinke Ren$^{123\dagger}$} \\
$^1$FNii-Shenzhen, $^2$SSE, and $^3$Guangdong Provincial Key Laboratory of Future Networks of Intelligence,\\ The Chinese University of Hong Kong, Shenzhen, China\\
$^4$ Harbin Engineering University, China \\
$^5$Infused Synapse AI, \href{www.isai.net.cn}{www.isai.net.cn}\\ 
\texttt{\small jinkeren@cuhk.edu.cn; hanyatong@cuhk.edu.cn}
}

\makeatletter
\AtBeginDocument{\let\c@lstlisting\c@figure}
\makeatother

\begin{document}

\maketitle

\nnfootnote{$\dagger$ Correspondence authors: Jinke Ren; Yatong Han.}
\nnfootnote{* These authors contributed equally to this work.}

\begin{abstract}
A key objective of embodied intelligence is enabling agents to perform long-horizon tasks in dynamic environments while maintaining robust decision-making and adaptability. To achieve this goal, we propose the Spatio-Temporal Memory Agent (STMA), a novel framework designed to enhance task planning and execution by integrating spatio-temporal memory. STMA is built upon three critical components: (1) a spatio-temporal memory module that captures historical and environmental changes in real time, (2) a dynamic knowledge graph that facilitates adaptive spatial reasoning, and (3) a planner-critic mechanism that iteratively refines task strategies. We evaluate STMA in the TextWorld environment on 32 tasks, involving multi-step planning and exploration under varying levels of complexity. Experimental results demonstrate that STMA achieves a 31.25\% improvement in success rate and a 24.7\% increase in average score compared to the state-of-the-art model. The results highlight the effectiveness of spatio-temporal memory in advancing the memory capabilities of embodied agents. 
\end{abstract}

\section{Introduction}
\label{sec:intro}
In embodied intelligence, the ability of agents to perform complex tasks in dynamic environments depends on their capabilities for long-term planning, reasoning, and adaptability. Despite recent advances in Artificial Intelligence (AI) systems powered by Large Language Models (LLMs), 
such as open-world games \cite{yan2023larp}, dialogue systems \cite{yi2024survey}, and personalized recommendation systems \cite{lyu2023llm}, there remain significant challenges in their applications to embodied intelligence \cite{duan2022survey, pfeifer2004embodied}. In particular, limited memory capacity prevents agents from effectively integrating historical information and adapting to evolving environments, resulting in reduced decision-making accuracy and poor task execution in long-horizon scenarios \cite{hatalis2023memory, fang2019scene}.

In recent years, researchers have explored various approaches to address this issue, such as Reinforcement Learning (RL) with memory augmentation \cite{parisotto2017neural}, dynamic memory networks \cite {kumar2021prediction}, and task reasoning using Knowledge Graphs (KGs) \cite{zhou2020graph}. While these methods have achieved notable successes, they often struggle to jointly model spatial and temporal information, which are essential for reasoning and planning in dynamic environments. Furthermore, existing approaches typically rely on fixed plans or limited context, which hinders adaptability and robustness in multi-task scenarios.

\begin{figure}[t]
\centering
   \includegraphics[width=0.76\linewidth]{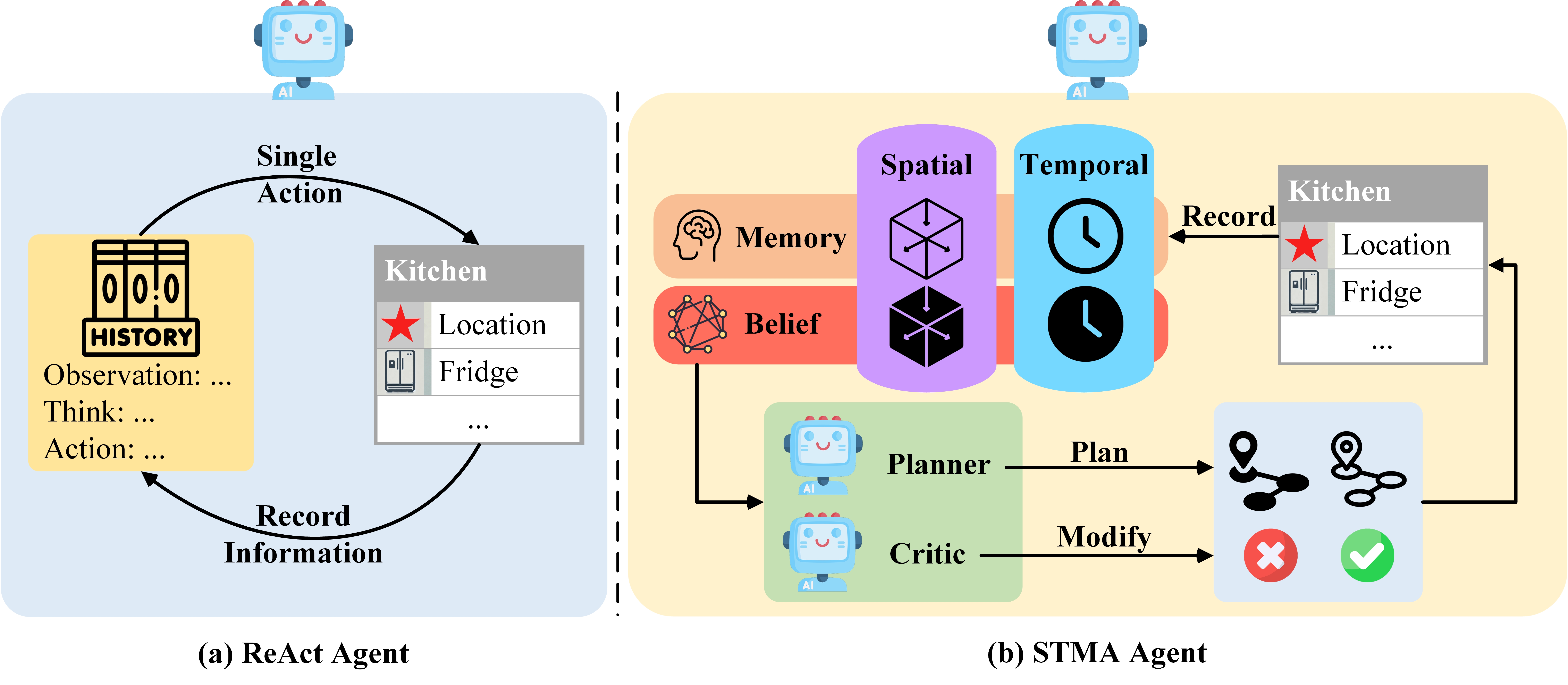}
   \caption{ {\textbf{Comparative overview of ReAct and STMA.}} {\it (a) ReAct} uses a simple history buffer to store action-feedback pairs and reasoning information, generating actions one step at a time. This approach lacks structured spatio-temporal reasoning, limiting its adaptability in complex, long-horizon tasks. {\it{(b) STMA}} utilizes dedicated spatial memory and temporal memory, summarized into refined spatial belief and temporal belief using the large model's capabilities. The planner-critic module enables closed-loop planning, dynamically validating and adjusting action sequences based on environmental feedback.} 
   \label{fig:intro}
\end{figure}

To address these issues, we introduce Spatio-Temporal Memory Agent (STMA), a new framework designed to enhance task planning and decision-making by integrating both temporal and spatial memory. This memory design reduces reliance on large models by enabling the agent to dynamically reflect on and correct suboptimal strategies, thus rapidly converging toward optimal solutions. 
Figure \ref{fig:intro} shows the comparison of the conventional framework ReAct \cite{yao2022react} and our proposed framework STMA. ReAct utilizes a simple history buffer to record action-feedback pairs along with some reasoning information, from which the next action is derived. In contrast, STMA replaces the traditional history buffer with spatio-temporal memory, significantly improving the agent's spatial reasoning capabilities. Furthermore, rather than directly using raw spatio-temporal memory, we capitalize on the summarization to distill the memory into a refined belief, allowing for more accurate decision-making by the planner based on this enriched input. Finally, we integrate the planner-critic, a closed-loop planning architecture, to enable the large model to generate a sequence of actions simultaneously, optimizing its planning capabilities, while dynamic feedback from the critic further enhances the plan's accuracy. 

We evaluate the performance of STMA in the Textworld environment \cite{cote2019textworld}, a partially observable, multi-task setting that requires agents to complete cooking tasks with varying complexities. 

Experimental results demonstrate that STMA outperforms the state-of-the-art models across several key metrics. Notably, STMA achieves a 31.25\% improvement in task success rate relative to the best-performing baseline, and a 24.7\% increase in average score, underscoring its enhanced ability to plan and adapt in complex, dynamic environments. 
The main contributions of this paper can be summarized as follows. \vspace{-1em}
\begin{itemize} \itemsep -3pt
    \item \textbf{Dynamic KG-based Spatial Memory}: We propose a dynamic KG for spatial memory that updates in real-time to reflect environmental changes. This approach enhances task reasoning and adaptability in complex, dynamic environments, improving long-horizon task planning over static memory systems. 
    \item \textbf{Planner-Critic Closed-Loop Architecture}: We present a planner-critic framework that combines proactive planning with real-time feedback. The planner generates multi-step action plans based on refined beliefs, while the critic adjusts strategies before each action step. This closed-loop mechanism ensures robust decision-making, particularly in environments with frequent re-planning and task switching. 
    \item \textbf{Spatio-Temporal Memory with Open-Source Model Efficacy}: Our agent integrates spatio-temporal memory within a closed-loop system, enabling dynamic adaptation to complex tasks. Remarkably, the agent performs competitively using open-source models (e.g., Qwen2.5-72b) without fine-tuning. Unlike architectures that rely heavily on large model capabilities, our approach demonstrates that a well-designed spatio-temporal memory system, combined with a planner-critic framework, can leverage open-source models for high-performance task execution.
\end{itemize}

\section{Related Work}
\label{sec:relate_work} The construction of embodied agents can be categorized into three approaches, including rule-based and traditional planning methods \cite{blake2001rule}, RL methods \cite{russell2003q}, and LLM-based methods \cite{zhao2024expel}. 
\vspace{-1em}
\paragraph{Rule-based and Traditional Planning Methods:} Rule-based and traditional planning methods rely on expert-designed rules or algorithms to solve tasks, making them suitable for simple tasks in static environments. For example, the A* algorithm \cite{liu2011comparative} excels in path planning but suffers from lower efficiency and accuracy in dynamic or unstructured environments \cite{mohanan2018survey}. Model-based planning \cite{ghallab2004automated} and constraint-based planning \cite{baptiste2006constraint} attempt to improve adaptability to complex environments, but challenges persist when handling uncertainty and dynamic changes.
\vspace{-1em}
\paragraph{RL Methods:} RL enables agents to learn task strategies through interaction with the environment and has achieved notable success in various domains, such as AlphaZero \cite{silver2018general} in board games. However, RL often faces issues in slow convergence and policy instability in high-dimensional action spaces and sparse reward scenarios \cite{vecerik2017leveraging, horgan2018distributed}. Additionally, RL performs poorly in tasks requiring global planning and long-term memory due to its focus on short-term returns \cite{nguyen2020deep, rusu2016progressive}. To address these challenges, researchers have proposed model-based RL \cite{kaiser2019model, moerland2023model} to improve sample efficiency by constructing environment models, but these methods still struggle with model complexity and generalization. Deep RL \cite{mnih2015human} utilizes neural networks to enhance state representation but faces computational and data efficiency challenges in multi-task and high-dimensional scenarios \cite{li2019deep, zhu2021overview}. Furthermore, hierarchical RL \cite{pateria2021hierarchical} and meta-RL \cite{beck2023survey, finn2017model} offer new approaches to handle long-term planning and task transfer, demonstrating potential in dynamic and complex environments. Despite these advances, RL still struggles with stability, computational overhead, and sample efficiency, particularly in the context of complex embodied tasks.
\vspace{-1em}
\paragraph{LLM-based Methods:} The introduction of LLMs has significantly enhanced the reasoning and task-handling capabilities of agents \cite{chan2023chateval, kannan2024smart, sun2024llm}. LLMs like GPT \cite{hurst2024gpt} and Gato \cite{reed2022generalist} leverage self-supervised learning to process multimodal data and excel in natural language understanding and open-world tasks \cite{raffel2020exploring}. However, existing LLM-driven agents exhibit limitations in long-term planning and dynamic task environments, manifesting two key issues---(1) Memory limitations: LLMs rely on autoregressive generation models and are unable to track task context or effectively store historical information; (2) Spatio-temporal reasoning deficits: LLMs perform reasoning based on pattern matching, lacking the ability to model spatio-temporal relationships in dynamic environments.

Recently, researchers have proposed several approaches to address these issues. 
For example, ReAct \cite{yao2022react} enhances task planning by introducing reflective and multi-step reasoning. However, ReAct's reasoning process relies on manually set few-shot samples, which limits its generalization. Reflexion \cite{shinn2024reflexion}, building upon ReAct, incorporates a self-reflection mechanism that allows agents to accumulate experience through multi-step trial and error. However, in embodied environments, errors may not be recoverable, limiting the effectiveness of this trial-and-error learning. Swiftsage \cite{lin2024swiftsage}, inspired by human dual-process theory \cite{frankish2010dual} and fast-slow thinking \cite{kahneman2011thinking}, combines these modules to handle complex tasks. However, its open-loop architecture fails to adequately support long-term memory and dynamic planning. AdaPlanner \cite{sun2024adaplanner} proposes a closed-loop architecture where an initial plan is refined based on environmental feedback. Nevertheless, it lacks a memory system, limiting its adaptability to long-horizon planning tasks. Hippo RAG \cite{gutierrez2024hipporag} mimics the human hippocampus \cite{burgess2002human} and introduces KGs as long-term memory indices \cite{chen2020review}, significantly enhancing knowledge retrieval. However, these methods are still confined to short-term reasoning and lack support for long-term planning in dynamic environments.

\section{Problem Formulation}
\label{sec:problem_formulation}
We model the agent-environment interaction as a Partially Observable Markov Decision Process (POMDP) \cite{kaelbling1998planning} extended with spatio-temporal reasoning. The POMDP is defined by the tuple \((\mathcal{S}, \mathcal{A}, \Omega, \mathcal{T}, \mathcal{O})\), where \(\mathcal{S}\) is the state space, \(\mathcal{A}\) the action space, \(\Omega\) the observation space, \(\mathcal{T}(s'|s,a)\) the transition dynamics, and \(\mathcal{O}(o|s',a)\) the observation function. 

At time-step \(i\), the agent receives observation \(o_i \sim \mathcal{O}(\cdot|s_i,a_{i-1})\) and maintains a belief \(b_i \in \Delta(\mathcal{S})\), a probability distribution over states conditioned on the interaction history:
\begin{equation}
    b_i(s) = P(s_i = s | o_{1:i}, a_{1:i-1}).
\end{equation}
Following the predictive processing theory \cite{kaelbling1998planning}, this belief serves as the agent's internal approximation of the true environmental state. In our work, we decompose the belief into:
\begin{equation}
    b_i = (b_i^t, b_i^s),
\end{equation}
where \(b_i^t\) captures temporal dependencies through historical action-observation sequences, and \(b_i^s\) encodes spatial relationships as a dynamic KG. The agent uses these factored beliefs to approximate both the temporal progression of states and the spatial configuration of objects, enabling joint reasoning about environmental dynamics. This formulation provides a theoretical foundation for integrating memory-augmented LLM agents with spatio-temporal awareness.

\begin{figure}[t]
\centering
   \includegraphics[width=1\linewidth]{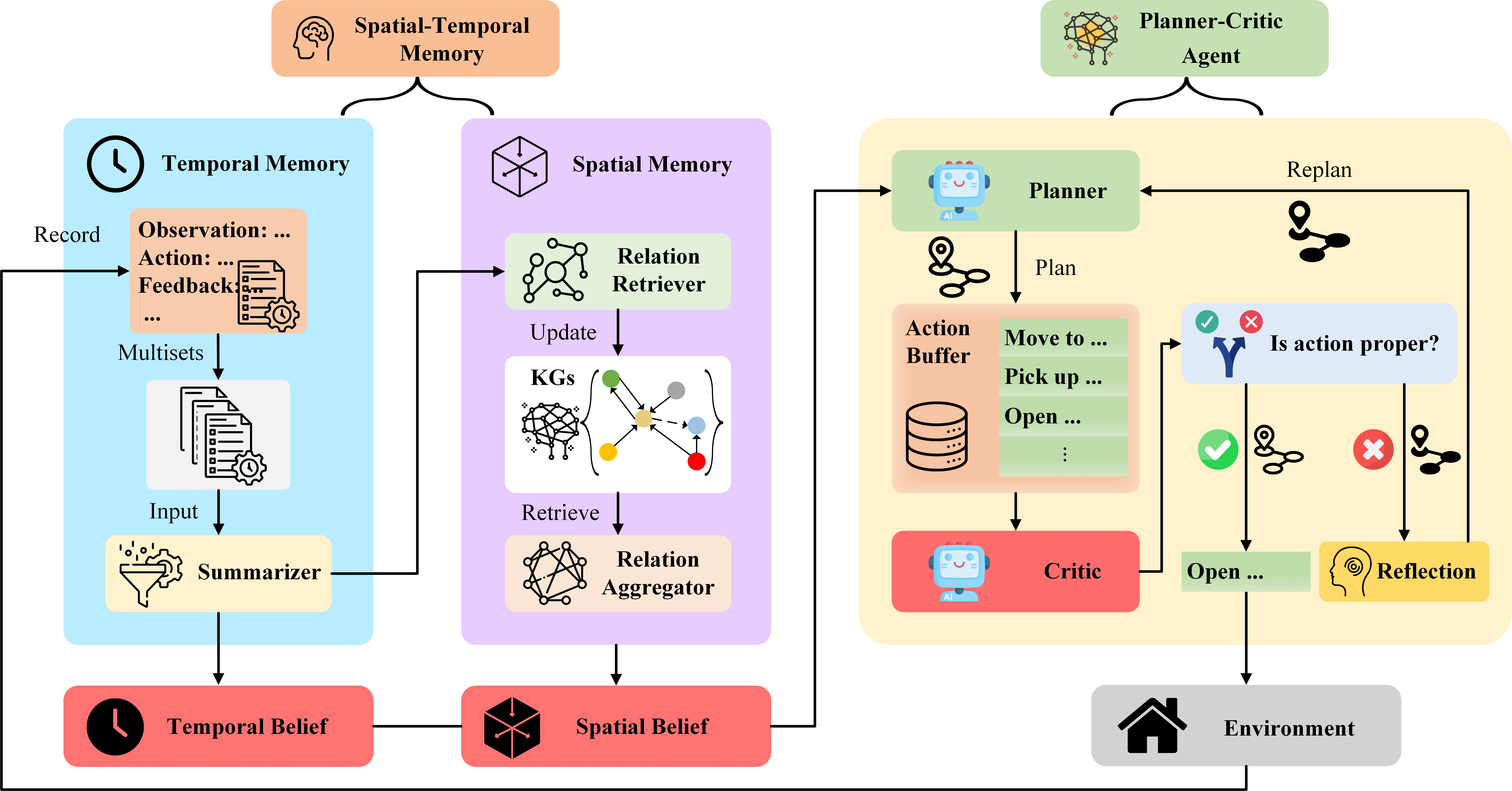}
   \caption{\textbf{Overview of STMA.} STMA consists of two components: a spatio-temporal memory module and a planner-critic module. The spatio-temporal memory module is divided into a temporal memory submodule and a spatial memory submodule, which provide temporal and spatial beliefs, respectively. These beliefs serve as the spatio-temporal context for the planner-critic module. The planner-critic module consists of a planner and a critic. The planner performs action planning based on the belief and generates multi-step plans in a single pass. The critic evaluates the plan before each action step, verifying whether the plan is correct and aligns with the most current environmental conditions.} 
   \label{fig:onecol}
\end{figure}

\section{Robotic Spatio-Temporal Memory Agent}
\label{sec:methodology}
As illustrated in Figure \ref{fig:onecol}, STMA is composed of three key components, including temporal memory, spatial memory, and a planner-critic module.

\subsection{Temporal Memory}
\label{sec:temporal_memory} 
Temporal memory maintains sequential event records to support time-dependent reasoning under partial observability. It addresses two key challenges: (1) LLMs struggle with long, unstructured histories, and (2) raw observations contain redundant details irrelevant to current decisions. Our solution combines raw data preservation with adaptive compression through two components:

\noindent\textbf{History Buffer.} 
A first-in-first-out queue storing raw interaction tuples is defined as:
\begin{equation}
\mathcal{H} = \{h_i\}_{i=1}^n,
\end{equation}
where $h_i = (o_i, a_i)$ preserves the complete interaction history. This ensures no information loss while providing temporal grounding for downstream processing.

\noindent\textbf{Summarizer.}  
Directly feeding raw interaction sequences to LLMs risks overwhelming them with redundant details and disordered temporal information. Our summarizer mitigates this via three key operations---(1) \textit{Information Compression}: Condenses raw observations $h_{[1:i-1]}$ into concise summaries, preserving task-relevant details while eliminating clutter. (2) \textit{Temporal Structuring}: Organizes historical data into standardized formats (e.g., completed actions, active goals, observations) to enhance prompt clarity. (3) \textit{Preliminary Reasoning}: Infers implicit patterns like progress tracking (retrieved key from the drawer) and strategic suggestions (consider examining the locked cabinet next).

This process converts raw histories into temporal belief states, which is given by:
\begin{equation}
    \mathcal{Sum}: h_{[1:i-1]} \rightarrow b_i^t,
\end{equation}
Where $b_i^t$ represents the temporal belief that compresses the temporal information for STMA. By maintaining compressed yet information-rich belief states, the summarizer enables efficient reasoning while retaining critical historical dependencies---striking a balance between memory efficiency and decision-making fidelity.

\subsection{Spatial Memory}
\label{sec:spatial_memory}
Spatial memory aims to provide the LLM with direct spatial information to address the model's limitations in spatial reasoning. By incorporating spatial memory, LLM can process and reason about spatial relationships, thereby improving its performance in tasks requiring spatial awareness and decision-making.

As shown in Figure \ref{fig:onecol}, the spatial memory is composed of four modules, including a relation retriever, a KG, a retrieve algorithm, and a relation aggregator. 

\noindent\textbf{Relation Retriever.}  
Inspired by HippoRAG \cite{gutierrez2024hipporag}, the relation retriever module leverages LLMs to derive spatial relationships from the temporal belief \(b^t_i\), which encapsulates compressed historical information. This module extracts structured representations of spatial interactions as a set of semantic triples:
\begin{equation}
G' = \big\{r_i \,|\, r_i = (x^s_i, x^r_i, x^o_i)\big\},
\end{equation}
where \(x^s_i\) denotes the subject entity, \(x^r_i\) the relationship type, and \(x^o_i\) the object entity. Formally, the retriever operates as a mapping function:
\begin{equation}
    R: b^t_i \rightarrow G',
\end{equation}
which enables the translation of temporally condensed beliefs into explicit spatial relational knowledge. This process bridges temporal reasoning with spatial awareness, which is critical for complex environment navigation.

\noindent\textbf{Dynamic KG.}  
To enhance system scalability and maintain stable long-term spatial memory, we formalize spatial relationships as a dynamic KG. The graph \(\mathcal{G}(V, E)\) comprises vertices \(V\) (entities) and edges \(E\) (relationships), updated iteratively as the agent interacts with the environment. At each step, newly extracted relationships \(G'\) replace outdated edges in \(G\), ensuring the graph reflects current spatial configurations. This dynamic update mechanism allows the agent to adapt to environmental changes while mitigating memory staleness. By decoupling transient observations from persistent relational structures, the KG provides a robust foundation for spatial reasoning and plan refinement.

\noindent\textbf{Retrieve Algorithm.} To extract task-relevant subgraphs from the KG while balancing semantic relevance and relational diversity, we propose a two-step retrieval process---(1) \textit{Semantic Filtering}: Compute cosine similarity between the query embedding (derived from task/environment context) and entity embeddings, retaining the top-$n$ entities. (2) \textit{Relational Expansion}: Perform $K$-hop neighborhood search from the filtered entities to capture local relational structures, forming the final subgraph.

This approach ensures the retrieved subgraph $\mathcal{G}_s$ preserves both semantic alignment with the current context and spatial relationships critical for planning. The full procedure is formalized in Algorithm~\ref{alg:retrieve algorithm}.

\begin{algorithm}[t]
\caption{Retrieve Algorithm}
\label{alg:retrieve algorithm}
\begin{algorithmic}[1]
\REQUIRE Query string $q$, hop count $K$, top-$n$ threshold, entity embeddings $\{\mathbf{e}_v\}_{v\in V}$.
\ENSURE Relevant subgraph $\mathcal{G}_s$.
\STATE Compute query embedding: $\mathbf{q} \gets \text{Embed}(q)$;
\STATE Initialize similarity scores: $Scores \gets \{\text{sim}(\mathbf{q}, \mathbf{e}_v) | v \in V\}$;
\STATE Select top-$n$ entities: $V_{\text{top}} \gets \text{argtop}_n(Scores)$; 
\STATE Initialize candidate set: $V_{\text{cand}} \gets V_{\text{top}}$;
\FOR{$k = 1$ \textbf{to} $K$}
    \STATE Expand neighbors: $V_{\text{cand}} \gets V_{\text{cand}} \cup \{u | (v,u) \in E, v \in V_{\text{cand}}\}$;
\ENDFOR
\STATE Extract subgraph: $\mathcal{G}_s \gets (V_{\text{cand}}, E_{\text{cand}})$ where $E_{\text{cand}} = \{(v,u) \in E | v,u \in V_{\text{cand}}\}$;
\STATE \textbf{return} $\mathcal{G}_s$.
\end{algorithmic}
\end{algorithm}

\noindent\textbf{Relation Aggregator.} Directly inserting the extracted list of triples as a string into the LLM prompt may degrade model performance. To address this issue, we introduce a relation aggregator to organize the list of triples into a natural language format. Similar to the summarizer, this module not only performs formatting but also incorporates a degree of spatial reasoning. For example, if \(A\) is west of \(B\) and \(B\) is west of \(C\), the module deduces that \(A\) is west of \(C\). This allows the relation aggregator to preprocess the extracted spatial memory, enabling reasoning over spatial relationships. The processed memory is then represented as the spatial belief \(b_t^s\). The relation aggregator is expressed as
\begin{equation}
    A: \mathcal{G}_s \rightarrow b_t^s,
\end{equation}
where \(\mathcal{G}_s\) is the task-relevant subgraph extracted by the retrieve algorithm.

\subsection{Planner-Critic Agent}
\label{sec:planner-critic}
The planner-critic module integrates spatio-temporal memory with systematic reasoning to generate robust action plans. As shown in Figure \ref{fig:onecol}, it operates via two cooperative components:

\noindent\textbf{Planner.}  
At step \(i\), the planner synthesizes the temporal belief \(b_i^t\), spatial belief \(b_i^s\), and current observation \(o_i\) to generate a subgoal \(g_i\) and corresponding action sequence \(\{\hat{a}_{i:k}\}_{k=1}^m\). Formally:
\begin{equation}
    P(b_i^t, b_i^s, o_i) \rightarrow (g_i, \{\hat{a}_{i:k}\}_{k=1}^m).
\end{equation}
This dual-belief integration enables coherent planning by contextualizing current observations within historical patterns and spatial constraints. The planner is implemented as a single agent that employs Chain-of-Thought (CoT) prompting \cite{wei2022chain} to explicitly reason about the current environment state, task requirements, and memory-derived constraints before outputting structured plans. Moreover, we urge a planner to think about the consequences of the actions it plans to do before it decides on the output.

\noindent\textbf{Critic.}  
Before executing each action \(\hat{a}_{j} \) at step \(j (j \in [i, k])\), the critic evaluates its validity using: (1) Temporal consistency with \(b_j^t\), (2) Spatial feasibility per \(b_j^s\), (3) Alignment with \(o_j\), and (4) Adherence to safety constraints. The evaluation function is given by:
\begin{equation}
    C(\hat{a}_{j}, b_j^t, b_j^s, o_i) \rightarrow (p_j, f_j),
\end{equation}
which outputs validity flag \(p_j \in \{\text{true}, \text{false}\}\) and feedback \(f_j\). If \(p_j = \text{false}\), the planner regenerates \(\{\hat{a}_{j:k}\}\) regarding \(f_j\), creating an iterative refinement loop. This closed-loop process mitigates hallucinations by grounding decisions in spatio-temporal reality.

Similar to the planner, the critic is implemented by a single CoT LLM agent. We ask the critic to think of the possible consequences of the proposed action, decide whether this action aligns with the subgoal planner provided in the current stage, and check whether this action is out-of-date as the environment changes. This CoT process will improve the robustness of STMA in dynamic environments.

The tight integration of memory-augmented beliefs with plan-critique cycles enables robust decision-making in partially observable environments. In the appendix \ref{sec:appendix_alg}, we formalize this procedure by  Algorithm \ref{alg:main_alg}, which demonstrates how spatio-temporal reasoning and iterative verification enhance plan reliability.

\section{Evaluation}
\label{sec:evaluation}

\begin{figure}
\centering
   \includegraphics[width=0.76\linewidth]{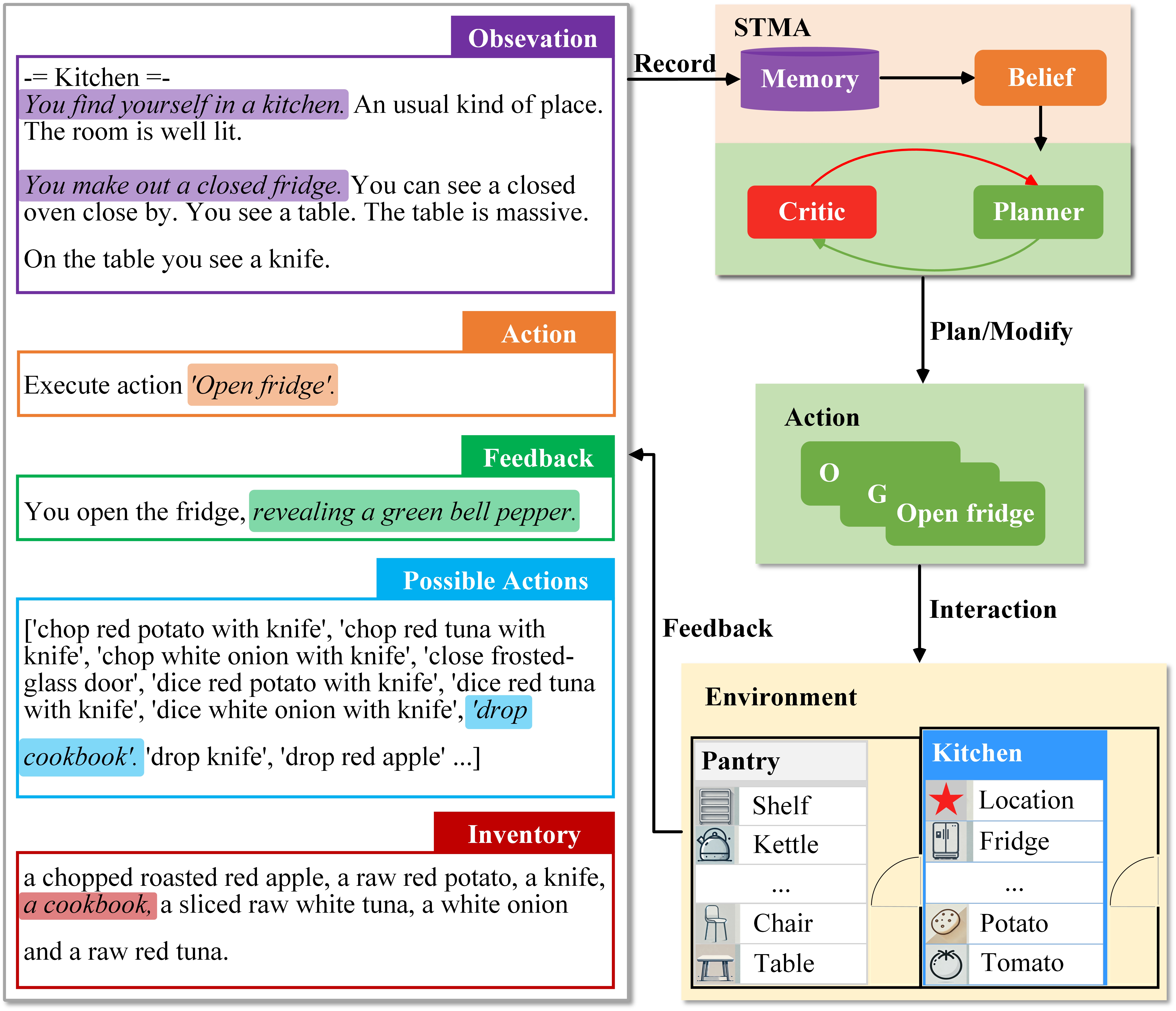}
   \caption{\noindent\textbf{Interaction with the Textworld Environment.} The interaction pattern between Textworld and our framework involves the environment providing the agent with the current observation, inventory, and a list of possible actions. Based on the agent's executed actions, the environment returns feedback. These pieces of information are recorded in STMA's spatio-temporal memory, serving as the necessary context for the planner-critic agent. Within this framework, the planner and critic collaborate to generate action plans and interact with the environment.}
   \label{fig:formulation}
\end{figure}

\subsection{Experiment Setting}
\label{sec:experiment_setting}
\noindent\textbf{Environments.} We use Textworld \cite{cote2019textworld} as our experimental environment. In Textworld, we design a series of cooking tasks with four difficulty levels. In these tasks, the agent is randomly placed in an indoor environment consisting of multiple rooms. It must explore the environment to locate the kitchen and complete the assigned cooking task. Upon reaching the kitchen, the agent follows the recipe's instructions to find the required ingredients and corresponding utensils within the environment. Note that the ingredients and utensils may not be located in the kitchen. Other evaluation details are provided in the appendix \ref{sec:exp_setting}.

The task difficulty is controlled by adjusting the number of rooms, the quantity of ingredients the agent needs to find, and the complexity of the recipe steps. Due to the unknown environment and the long sequences of actions required, this setting demands strong spatio-temporal reasoning skills to locate items and robust planning capabilities to execute the recipe instructions in the correct order using the appropriate utensils. Additionally, as the room layout and the distribution of items are unknown, the environment also requires the agent to exhibit high robustness in handling such uncertainties.

\noindent\textbf{Baselines.} To facilitate comparisons, we consider three baseline agent frameworks: (1) ReAct \cite{yao2022react}, which is an early framework designed for agent operations. It provides a prompting method that allows large models to exhibit reasoning capabilities while interacting with the environment through actions; (2) Reflexion \cite{shinn2024reflexion}, which introduces long-term memory and a self-reflection module. By employing a trial-and-error approach, Reflexion uses the self-reflection module to summarize experiences as long-term memory, thereby enhancing the model's capabilities; and (3) AdaPlanner \cite{sun2024adaplanner}, which proposes a closed-loop agent architecture. This framework continuously refines its plan to handle unforeseen situations, enabling dynamic adjustments to its strategy.


\begin{table}[t]
\caption{Comparison of Success Rates (SR) and Average Scores (AS) across difficulty levels between STMA and baseline methods. Performance metrics are reported as \(AS \pm \sigma\), where \(\sigma\) denotes the standard deviation across different difficulty levels.}
\label{tab:baseline}
\begin{centering}
\resizebox{\textwidth}{!}{
\begin{tabular}{lcccccccc}
\toprule
\multicolumn{1}{l}{Difficulty Level} & 
\multicolumn{2}{c}{1} & 
\multicolumn{2}{c}{2} & 
\multicolumn{2}{c}{3} & 
\multicolumn{2}{c}{4} \\
\cmidrule(lr){2-3} \cmidrule(lr){4-5} \cmidrule(lr){6-7} \cmidrule(lr){8-9}
 & SR & AS & SR & AS & SR & AS & SR & AS \\
\midrule
ReAct(GPT-4o) & 100.0 & 100.0$\pm$0.0 & 62.5 & 76.5$\pm$31.9 & 50.0 & 65.0$\pm$39.4 & 25.0 & 43.3$\pm$37.9 \\
Reflexion(GPT-4o) & 100.0 & 100.0$\pm$0.0 & 62.5 & 78.6$\pm$30.3 & 50.0 & 60.0$\pm$40.9 & 25.0 & 51.0$\pm$31.9 \\
AdaPlanner(GPT-4o) & 87.5 &90.6$\pm$24.8 & 37.5 & 51.8$\pm$43.4 & 0.0 & 3.8$\pm$4.8 & 0.0 & 1.9$\pm$3.3 \\
\textbf{STMA(GPT-4o)} & \textbf{100.0} & \textbf{100.0$\pm$0.0} & \textbf{75.0} & \textbf{91.1$\pm$15.9} & \textbf{75.0} & \textbf{87.5$\pm$26.3} & \textbf{37.5} & \textbf{60.6$\pm$38.5} \\
\midrule
ReAct(Qwen2.5-72b)  & 75.0 & 75.0$\pm$43.3 & 37.5 & 60.7$\pm$34.1 & 12.5 & 37.5$\pm$31.9 & 0.0 & 16.3$\pm$9.0 \\
Reflexion(Qwen2.5-72b) & 100.0 & 100.0$\pm$0.0 & 37.5 & 60.7$\pm$32.5 & 25.0 & 61.2$\pm$28.0 & 0.0 & 19.2$\pm$7.7 \\
AdaPlanner(Qwen2.5-72b) & 100.0 & 100.0$\pm$0.0 & 37.5 & 53.6$\pm$41.5 & 0.0 & 3.8$\pm$4.8 & 0.0 & 3.8$\pm$5.4 \\
\textbf{STMA(Qwen2.5-72b)}& \textbf{100.0} & \textbf{100.0$\pm$0.0} & \textbf{100.0} & \textbf{100.0$\pm$0.0} & \textbf{62.5} & \textbf{81.2$\pm$27.6} & \textbf{25.0} & \textbf{58.7$\pm$30.3} \\
\bottomrule
\end{tabular}
}
\end{centering}
\end{table}

In our experiments, each agent framework is tested using two LLMs: GPT-4o \cite{hurst2024gpt} and Qwen2.5-72b-instruct \cite{qwen2.5}. For embedding tasks, we uniformly use the nomic-embed-text-v1.5 model \cite{nussbaum2024nomic}. GPT-4o is one of the most commonly used proprietary LLMs, while Qwen2.5-72b-instruct represents a high-performing open-source alternative. Notably, Qwen2.5-72b-instruct demonstrates performance comparable to GPT-4o in several benchmark tasks \cite{white2024livebench}.

\noindent\textbf{Evaluation metrics.} We define two evaluation metrics to assess the performance: (1) Success Rate (SR), which is the ratio of completed tasks to the total number of tasks in each difficulty level. This metric reflects the agent's ability to complete tasks across randomly generated scenarios; (2) Average Score (AS), which is the ratio of intermediate scores achieved to the maximum possible score in each scenario. An AS of 100\% indicates that the task is completed in the given scenario.

\begin{figure}[t]
\centering
   \includegraphics[width=1.0\linewidth]{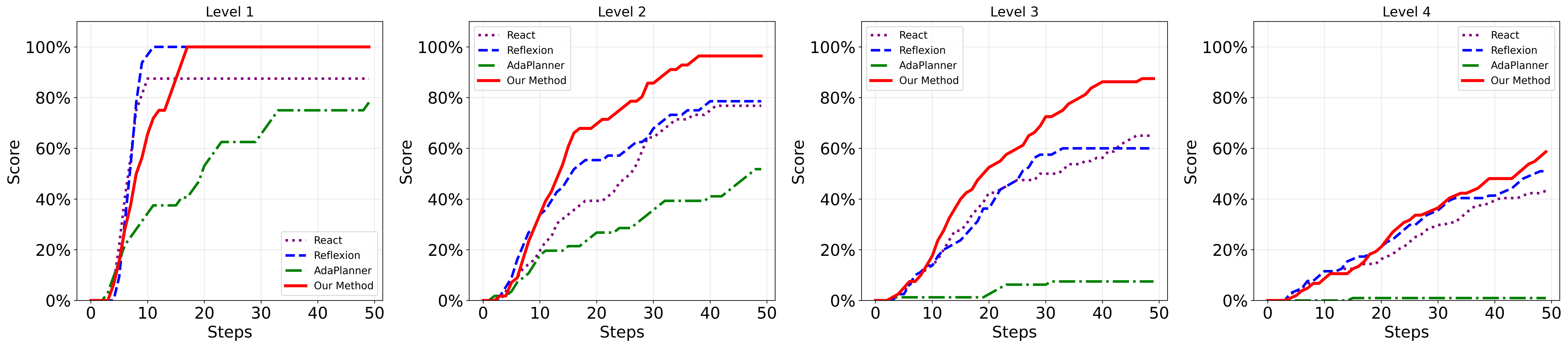}
   \caption{Average score vs. steps of different frameworks (powered by GPT-4o)}
   \label{fig:score_curve}
\end{figure}

\subsection{Performance Evaluation}
\label{sec:performance}

Table \ref{tab:baseline} shows the comparison of our agent framework with the three baseline frameworks across four difficulty levels. For each difficulty level, we report the AS across eight randomly generated scenarios, along with its standard deviation (represented as mean \(\pm\) standard deviation). Additionally, the SR for each difficulty level is recorded. 

For all models, AS and SR show a general downward trend as task difficulty increases. This reflects the effectiveness of our experimental design in creating a gradient of increasing complexity. As the environment becomes more challenging and tasks require more steps to complete, the SRs of all frameworks decrease. However, our framework consistently outperforms the baselines across all difficulty levels, regardless of the underlying model. The result stems from the integration of the spatio-temporal memory module, which enhances planning and execution capabilities. Specifically, our framework achieves an average improvement in SR of 12.5\% (GPT-4o powered) and 31.25\% (Qwen2.5-72b-instruct powered) compared to the best baseline. Similarly, the AS improves by an average of 11.15\% (GPT-4o powered) and 24.7\% (Qwen2.5-72b-instruct powered). Notably, the improvement is more pronounced for the open-source Qwen2.5-72b-instruct, demonstrating the framework’s ability to enhance the performance of less powerful models.

Compared to ReAct and Reflexion, STMA incorporates a spatio-temporal memory module and a planner-critic agent, which significantly enhance reasoning capabilities and improve the robustness of the agent through the critic mechanism. As a result, STMA demonstrates notable improvements in long-horizon planning tasks over ReAct. Besides, Reflexion's self-reflection mechanism proves effective in the Textworld cooking tasks, leading to a slight performance improvement over ReAct. However, experimental results indicate that this enhancement is relatively modest. While AdaPlanner also employs a critic-like mechanism, its reliance on Python code as the action space and a poorly designed memory system limits its adaptability to the Textworld cooking tasks, which require handling limited observations and long-term planning. In practice, AdaPlanner exhibits weaker spatial exploration capabilities than STMA.


Figure \ref{fig:score_curve} illustrates the AS progression curves for different models powered by GPT-4o. It is evident that our STMA framework consistently achieves the highest scores at a faster rate in most cases. This advantage stems from the collaboration between the planner and critic, which reduces suboptimal or erroneous actions generated by the planner.

For Level 1 tasks, the score progression of STMA is slightly lower than ReAct and Reflexion. It is because, in relatively simple tasks, the planner-critic agent in STMA behaves more cautiously, ensuring the correctness of certain actions before executing them. Additionally, GPT-4o-powered STMA tends to explore the environment more extensively before completing tasks. As a result, STMA exhibits a minor disadvantage in tasks with minimal spatial exploration requirements. However, for Level 2, 3, and 4 tasks, the exploration-oriented nature of STMA, combined with the enhanced spatial understanding enabled by its robust spatio-temporal memory framework, delivers significantly better performance. These results highlight STMA's superior ability to handle more complex tasks requiring advanced planning and spatial reasoning.

\begin{table}[t]
\caption{Ablation studies of the STMA. Success Rate (SR) and Average Score (AS) (\(AS \pm \sigma\)) across four difficulty levels for model variants with individual modules are removed.} 
\label{tab:ablation}
\begin{centering}
\resizebox{\textwidth}{!}{%
\begin{tabular}{lcccccccc}
\toprule
\multicolumn{1}{l}{Difficulty Level} & 
\multicolumn{2}{c}{1} & 
\multicolumn{2}{c}{2} & 
\multicolumn{2}{c}{3} & 
\multicolumn{2}{c}{4} \\
\cmidrule(lr){2-3} \cmidrule(lr){4-5} \cmidrule(lr){6-7} \cmidrule(lr){8-9}
 & SR & AS & SR & AS & SR & AS & SR & AS \\
\midrule
STMA w/o Spatio-Temporal Memory  & 0.0 & 0.0$\pm$0.0 & 0.0 & 0.0$\pm$0.0 & 0.0 & 0.0$\pm$0.0 &0.0 & 0.0$\pm$0.0 \\
STMA w/o Summarizer  & 87.5 & 87.5$\pm$33.1 & 100.0 & 100.0$\pm$0.0 & 25.0 & 55.0$\pm$32.8 & 12.5 & 49.0$\pm$28.5 \\
STMA w/o Summarizer for Spatial Memory  & 100.0 & 100.0$\pm$0.0 & 50.0 & 65.0$\pm$40.6 & 12.5 & 37.5$\pm$31.9 & 12.5 & 37.5$\pm$35.1 \\
STMA w/o Spatial Memory  & 87.5 & 93.8$\pm$16.5 & 75.0 & 91.1$\pm$15.9 & 50.0 & 65.0$\pm$40.6 & 12.5 & 30.8$\pm$28.0 \\
STMA w/o Relation Aggregator & 100.0 & 100.0$\pm$0.0 & 87.5 & 96.4$\pm$9.4 & 62.5 & 71.2$\pm$37.2 & 25.0 & 51.0$\pm$34.8 \\
STMA w/o Critic & 100.0 & 100.0$\pm$0.0 & 75.0 & 85.7$\pm$25.8 & 50.0 & 70.0$\pm$33.5 & 0.0 & 26.0$\pm$31.0 \\
\textbf{STMA}    & \textbf{100.0} & \textbf{100.0$\pm$0.0} & \textbf{100.0} & \textbf{100.0$\pm$0.0} & \textbf{62.5} & \textbf{81.2$\pm$27.6} & \textbf{25.0} & \textbf{58.7$\pm$30.3} \\
\bottomrule
\end{tabular}%
}
\end{centering}
\end{table}

\subsection{Ablation Studies}
\label{sec:ablation}

We conduct ablation studies on agents powered by Qwen2.5-72b-instruct. The results are presented in Table \ref{tab:ablation}.

\noindent\textbf{Spatio-Temporal Memory.} In our experiments, completely removing the spatio-temporal memory module makes the agent unable to complete any task. For tasks involving unknown environments and requiring long-term planning, memory is crucial. Without a memory system, the large model is incapable of ``remembering the recipe," making it impossible to complete even the simplest cooking tasks. Since our experimental setup requires the agent to recall and execute the steps specified in the recipe, the absence of memory renders task completion unattainable. Consequently, STMA without a memory system scores zero across all difficulty levels. This highlights the critical role of memory in POMDP-based systems, where constructing a cognitive understanding of the world state heavily relies on a robust memory framework.

\noindent\textbf{Summarizer.} The information in the history buffer is passed directly to the model as temporal belief in chronological order, without summarization (other components remain unchanged, with spatial memory still being extracted from summarized information). Results indicate that agent performance decreases slightly for simpler tasks (Levels 1 and 2) but drops significantly for more complex tasks. This suggests that for long-horizon tasks, the growing length of the history leads to an increasingly long prompt, which degrades the agent's performance. Additionally, the longer history dilutes the density of useful information, making the LLM more susceptible to irrelevant data. The summarizer mitigates this issue by condensing the temporal memory, reducing the adverse effects of long history as the number of steps increases. Therefore, the Summarizer is critical for STMA’s efficacy in complex scenarios, ensuring efficient information flow and robust decision-making by balancing detail with concision.

\noindent\textbf{Summarizer for Spatial Memory.} The information in the history buffer is passed directly to the spatial memory module in chronological order (while summarized information is still used as a temporal belief to control variables). Results show that agent performance decreases as task complexity increases, and the decline is more significant compared to removing the summarizer for temporal belief. This indicates that the process of generating spatial belief from spatial memory relies on the summarizer for initial preprocessing. This is due to the LLM's limited capability to extract relevant spatial information from an unprocessed lengthy history. Thus, we can observe that LLMs may not be good at implicitly summarizing temporal trajectories while completing other tasks based on the information that the trajectory contains. This limitation affects not only inference tasks such as planning but also information extract tasks such as relation retrieval. Therefore, the summarizer is essential for elevating the performance of downstream tasks.

\noindent\textbf{Spatial Memory.} We remove the spatial memory and its corresponding spatial belief input to evaluate the agent's reliance on temporal belief alone. Results show a significant performance drop, particularly for tasks with higher spatial complexity. This decline highlights the critical role of spatial memory in providing agents with a foundational ``sense of space," enabling them to build an internal ``mental map" of the environment during exploration. Notably, when comparing this experiment to the one where the summarizer for spatial memory is removed, we observe that eliminating spatial memory yields better performance across all but the most challenging Level 4 tasks. This suggests that agents require highly accurate spatial beliefs for effective decision-making. Incorrect or inaccurate spatial beliefs may mislead the agent, performing worse than when no spatial belief is provided. This underscores one of the key challenges in designing effective spatial memory systems. Thus, the spatial memory provides STMA with a ``sense of space," enabling it to make better decisions based on the spatial information.

\noindent\textbf{Relation Aggregator.} We remove the relation aggregator and use the raw list of relation triples extracted by the retrieve algorithm module as input without summarization or conversion into natural language. Results show a moderate performance decline compared to the baseline. While LLMs demonstrate some ability to interpret spatial relationships from a list of relation triples, converting these triples into natural language facilitates a better understanding of the spatial information encoded in the spatial belief. This indicates that while the relation aggregator is not strictly necessary, it significantly improves the LLM's ability to process and utilize spatial memory.

\noindent\textbf{Critic.} We remove the critic module from STMA, leaving only the planner to generate plans, which are executed without validation. Results show that while performance on the simplest tasks remains relatively stable, agent performance is significantly weakened for Levels 2, 3, and 4. This is attributed to the complexities of the cooking environment, which involves numerous dynamic and unpredictable elements. Without the critic’s involvement, any errors or hallucinations in the planner’s decisions are directly executed, leading to more redundant or incorrect actions in more complex tasks. For instance, during environment exploration, the agent often failed to update its plan upon encountering new information, continuing to follow outdated plans instead. This behavior results in inefficient exploration of new environments. Similarly, when executing recipe steps, errors such as using the wrong utensils are uncorrected, leading to task failures.

In addition, we observe that the LLM’s performance as a critic is generally stronger than its performance as a planner. This may be because the critic’s role is a classifier—determining whether an action is ``correct” or ``incorrect” based on beliefs and current environment. The classification task seems simpler than the planner’s generative task of creating new plans. This is one of the reasons why the critic is able to detect the error action generated by the planner. Therefore, removing the critic leads to a significant drop in performance, as the planner’s unchecked errors accumulate and negatively impact the agent’s overall effectiveness.

\section{Conclusion}
\label{sec:conclusion}
This paper proposes a new STMA framework for long-horizon embodied tasks, demonstrating the great potential of spatio-temporal memory mechanisms in task planning. Experimental results show that STMA performs exceptionally well across various task difficulty levels. Notably, when tackling complex tasks, STMA outperforms agents relying on random exploration by providing more precise and efficient solutions through rapid strategy adjustments and higher task completion rates.  It is noteworthy that even with open-source models like Qwen2.5-72b, STMA achieves performance comparable to proprietary models, validating the superiority of the spatio-temporal memory module. Future work may optimize the spatio-temporal memory module to enhance its adaptability in more complex tasks.

\bibliography{main} 
\bibliographystyle{unsrt}


\appendix

\section{Experiment Settings}
\label{sec:exp_setting}

\subsection{Environment Details}

\noindent\textbf{Task Settings.} We design four progressive difficulty levels with the following characteristics (summarized in Table~\ref{tab:task_settings}): Level 1 requires no ingredient search in a 6-room environment, while higher levels (2-4) introduce larger environments (9-12 rooms), more required ingredients (1-3), and longer recipes (6-10 steps). All tasks enforce a 50-turn limit to prioritize deliberate planning over random exploration.

\noindent\textbf{Difficulty Settings.} Environments feature three key constraints: (1) Ingredients require knife processing (cut/dice/slice) before cooking, (2) Three cooking methods (grill/fry/roast) require specific appliances (BBQ/stove/oven), (3) Randomized room layouts with door connections requiring explicit \textit{open} actions. Agent starting positions are randomized per scenario.

\noindent\textbf{Scoring Mechanism.} As detailed in Table~\ref{tab:task_settings}, maximum scores increase with difficulty (4-13 points). Agents earn 1 point per: ingredient collection, correct appliance usage, proper ingredient processing, and final meal preparation. Critical failures include: incorrect appliance/processing selection, repeated heat application causing burns, and turn limit expiration. Game rules are explicitly provided to agents via prompts to focus evaluation on strategic execution rather than knowledge acquisition.

\noindent\textbf{Common settings.} To prevent large models from obtaining scores through random action generation and to better evaluate their planning capabilities, we impose a constraint that each task must be completed within 50 turns. 

\begin{table}[t]
\label{tab:task_settings}
\centering
\resizebox{\textwidth}{!}{
\begin{tabular}{|c|c|c|c|c|c|}
\hline
\multicolumn{6}{|c|}{\textbf{Task Difficulty Settings}} \\
\hline
\textbf{Level} & \textbf{Rooms} & \textbf{Ingredients to Find} & \textbf{Steps in Recipe} & \textbf{Ingredients in Recipe} & \textbf{Total Score} \\
\hline
1 & 6 & 0 & 4 & 1 & 4 \\
\hline
2 & 9 & 1 & 6 & 2 & 7 \\
\hline
3 & 9 & 2 & 8 & 3 & 10 \\
\hline
4 & 12 & 3 & 10 & 4 & 13 \\
\hline
\multicolumn{6}{|c|}{\textbf{Scoring Points and Task Failures}} \\
\hline
\multicolumn{3}{|c|}{\textbf{Scoring Points}} & \multicolumn{3}{c|}{\textbf{Task Failures}} \\
\hline
\multicolumn{3}{|l|}{Collecting specified ingredients} & \multicolumn{3}{l|}{Using an incorrect cooking appliance} \\
\hline
\multicolumn{3}{|l|}{Processing ingredients with correct cooker} & \multicolumn{3}{l|}{Incorrect processing technique (cut, dice, slice)} \\
\hline
\multicolumn{3}{|l|}{Handling ingredients properly (cut, dice, slice)} & \multicolumn{3}{l|}{Repeating heat-based cooking steps causes burning} \\
\hline
\multicolumn{3}{|l|}{Executing final steps (e.g., "prepare meal")} & \multicolumn{3}{l|}{Exceeding 50 turns without task completion} \\
\hline
\end{tabular}
}
\caption{Task Difficulty Settings, Scoring Points, and Task Failures}
\end{table}

This configuration isolates evaluation to strategic reasoning through three control mechanisms: explicit rule provision, randomized spatial configurations, and time-constrained execution.

The sample instructions for constructing games in different difficulty levels are shown below:

Level 1:
\begin{lstlisting}
tw-make tw-cooking --recipe 1 --take 0 --go 6 --open --cook --cut --output ./game_0_1.ulx -f -v --seed 1001
\end{lstlisting}
Level 2:
\begin{lstlisting}
tw-make tw-cooking --recipe 2 --take 1 --go 9 --open --cook --cut --output ./game_1_1.ulx -f -v --seed 1001
\end{lstlisting}
Level 3:
\begin{lstlisting}
tw-make tw-cooking --recipe 3 --take 2 --go 9 --open --cook --cut --output ./game_2_2.ulx -f -v --seed 20002
\end{lstlisting}
Level 4:
\begin{lstlisting}
tw-make tw-cooking --recipe 4 --take 3 --go 12 --open --cook --cut --output ./game_3_3.ulx -f -v --seed 303
\end{lstlisting}

\subsection{Hyperparameters.} 

In our experiments, STMA utilizes three key hyperparameters:

\noindent\textbf{Number of vertices retrieved in the first step of the retrieval algorithm} In the first step of the retrieval process, we use cosine similarity to extract 8 vertices. This provides a sufficient initial set of vertices for the subsequent K-hop neighbor search.

\noindent\textbf{K value in the K-hop algorithm}: During experiments, we set \(k = 3\) for the K-hop algorithm. This choice is based on the complexity of the environment. Given that the task environment contains a maximum of 12 rooms, a depth of 3 ensures that the model can access at least three-hop neighbors, covering the majority of spatial relationships required for task completion.

\noindent\textbf{Record length in the history buffer:} We limit the history buffer to store records from the most recent 25 iterations. Since each task is constrained to a maximum of 50 iterations, maintaining a record of the last 25 iterations is sufficient and helps reduce the input length for the model. This not only minimizes memory usage but also lowers computational costs without compromising task performance.

\section{Additional Experiments}
\begin{figure}[p]
\centering
   \includegraphics[width=1.0\linewidth]{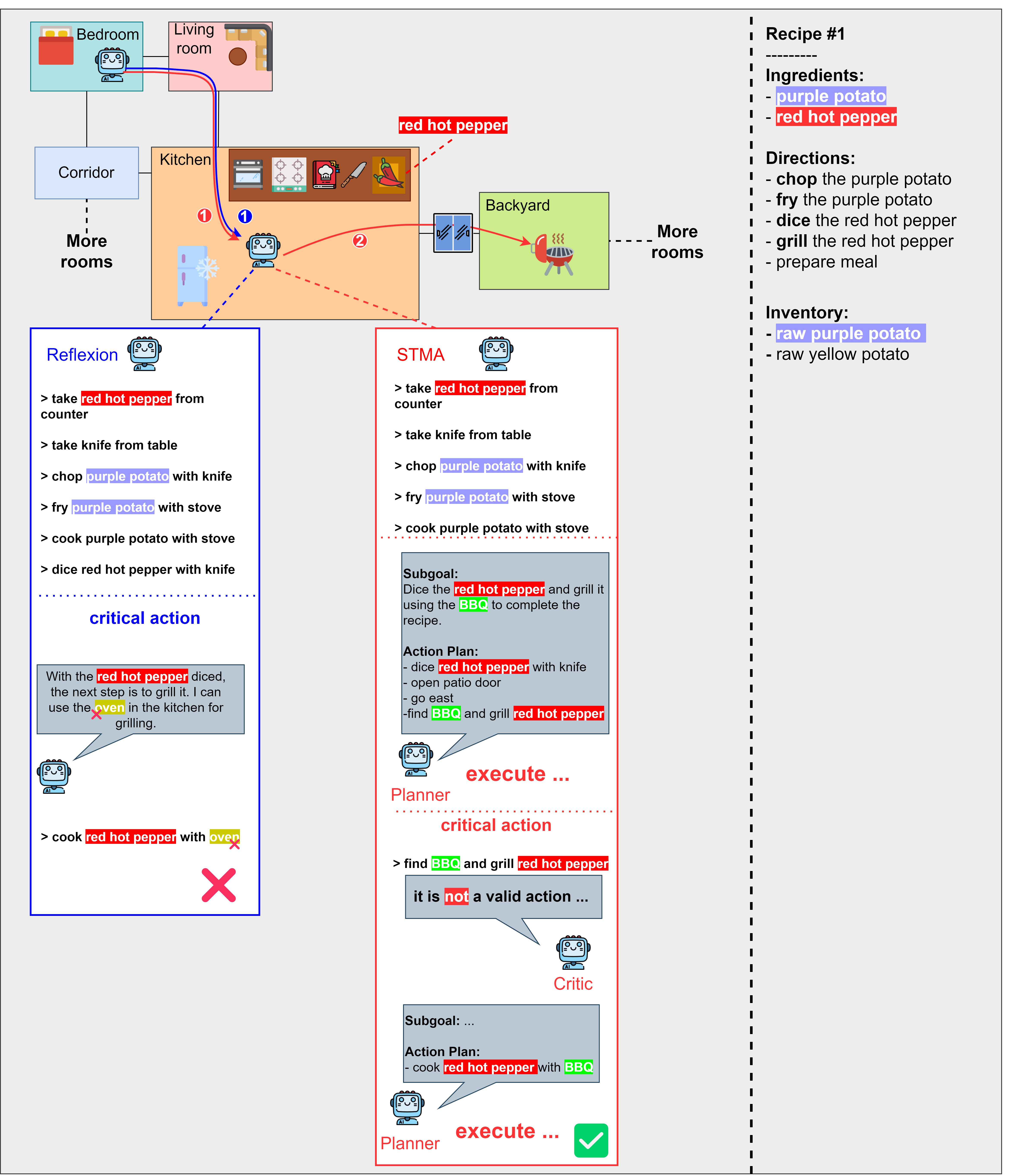}
   \caption{STMA versus Reflexion in Case 1.}
   \label{fig:1-1}
\end{figure}

\begin{figure}[p]
\centering
   \includegraphics[width=1.0\linewidth]{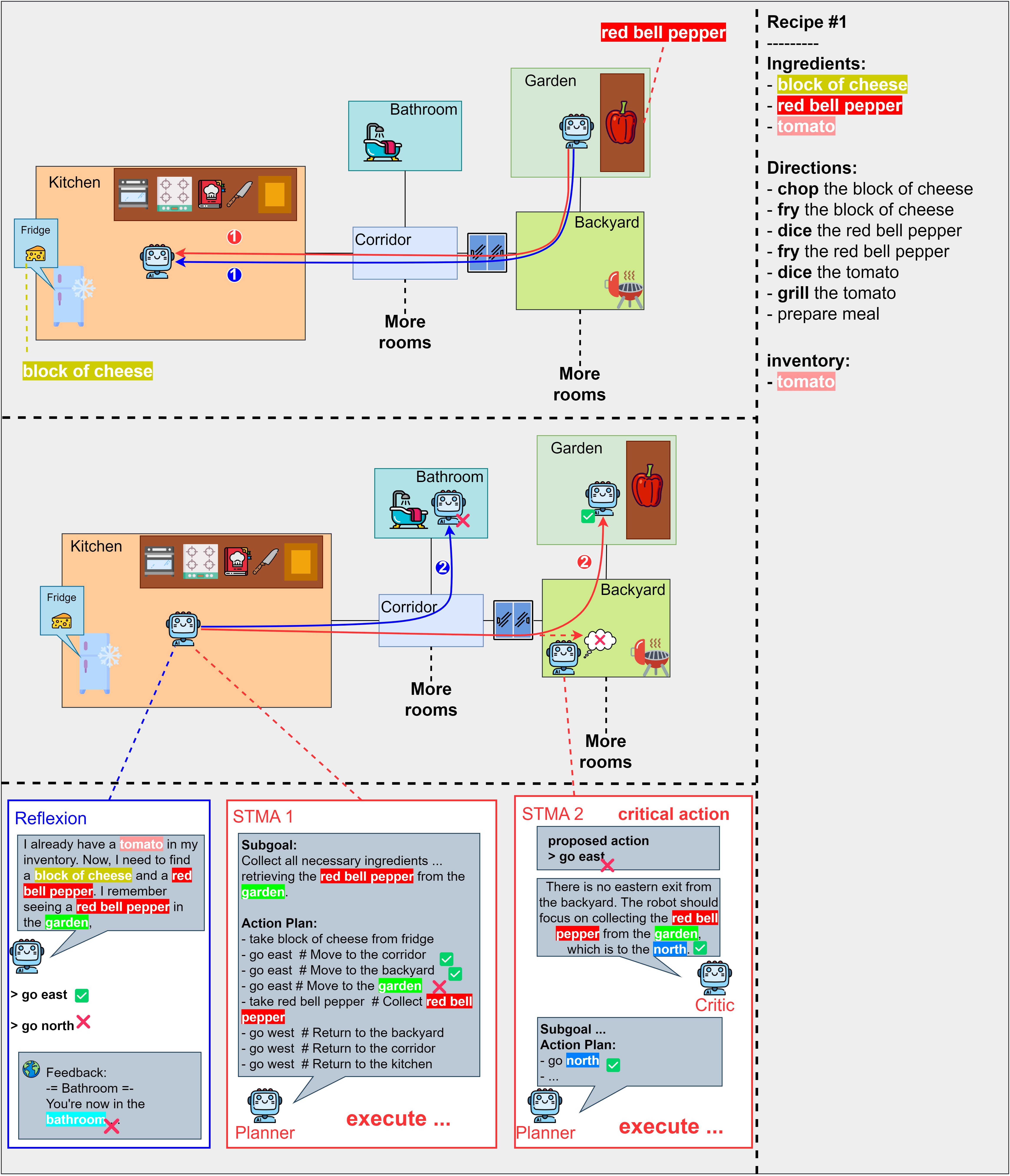}
   \caption{STMA versus Reflexion in Case 2.}
   \label{fig:2-8}
\end{figure}

To further illustrate the behavior of our agent in real embodied environments, we select two representative scenarios from all test cases. These scenarios are used to compare the performance of STMA with the best-performing baseline model in the same context. The goal is to highlight the advantages of STMA relative to other baselines within these scenarios. For this case study, all agents are powered by GPT-4o.

\subsection{Case 1: The Role of the Critic.}
\label{sec:case1}

Environment 1 represents a scenario classified under Difficulty Level 2. Among the baseline models evaluated in this setting, Reflexion achieved the highest performance. Consequently, we conducted a comparative analysis between Reflexion and the STMA agent, focusing on a representative subtask of this environment. The experimental results are illustrated in Figure \ref{fig:1-1}. In this scenario, both Reflexion and STMA successfully explored the kitchen environment and retrieved a recipe. The acquired recipe and the agent's initial inventory are displayed in the right panel of Figure \ref{fig:1-1}. Both agents executed operations based on the recipe instructions. However, Reflexion failed to complete the task due to selecting an incorrect cooker for ingredient processing at a critical step. In contrast, STMA recognized the necessity of using the appropriate cooker and generated a logically coherent plan. Notably, the final step of STMA's initial plan proposed by the Planner constituted an invalid action within the Textworld environment. Upon execution, the Critic module identified this discrepancy, prompting the Planner to revise its strategy. After reformulation, the Planner successfully generated a valid sequence of operations adhering to procedural constraints.

This experiment demonstrates that the collaborative framework between the Critic and Planner in STMA enables the agent to achieve superior long-term planning capabilities while maintaining robustness against errors caused by LLM hallucinations or inherent limitations. Compared to baseline models such as Reflexion, STMA exhibits enhanced planning reliability and adaptability, underscoring the efficacy of its self-corrective mechanism in dynamic task environments.

\subsection{Case 2: Demonstrating Collaborative Capabilities of Spatio-temporal Memory and Planner-Critic in the STMA Framework.}
\label{sec:case2}

Environment 2 represents a Difficulty Level 3 task designed to evaluate the agent's capacity to handle intricate operational workflows and manage a larger set of required ingredients. Similar to Environment 1, we compare the performance of the STMA framework against Reflexion, the strongest baseline framework. A representative subtask is illustrated in Figure \ref{fig:2-8}. In this environment, the agent is initialized in the  ``Garden" sub-environment, which contains the critical ingredient ``red bell pepper." However, since the recipe resides in the ``Kitchen," the agent initially lacks awareness of this ingredient's necessity. This configuration evaluates the agent's memory retention capability, requiring it to recall the Garden's relevance after discovering the recipe in the Kitchen.

As shown in the top-left panel of Figure \ref{fig:2-8}, both Reflexion and STMA successfully navigate to the Kitchen and retrieve the recipe, subsequently recognizing the need to return to the Garden. However, the second-left panel reveals a critical divergence: Due to the absence of spatial memory, Reflexion executes an erroneous decision, leading to the exploration of an irrelevant room. In contrast, STMA leverages its spatial memory to formulate an optimal path. Although the final planning step initially contains an error, the Critic module intervenes by analyzing the Spatial Belief state, enabling the Planner to self-correct and select the correct trajectory.

This case highlights three key observations regarding the STMA framework. 1) While spatial memory enhances the Planner's spatial reasoning, it does not fully mitigate erroneous planning—a limitation potentially attributable to large language model hallucinations. 2) The Critic module demonstrates advanced functionality beyond binary action validation, providing actionable suggestions for plan refinement based on Spatial Belief analysis. 3) For Reflexion, we note that despite it do not have spatial memory like STMA, with GPT-4o's capabilities, the framework exhibits inferior spatio-temporal reasoning compared to STMA. This performance gap underscores the critical role of explicit spatio-temporal memory integration in complex embodied agent tasks.

This case demonstrates the synergistic collaboration among core components within the STMA framework. The temporal memory module maintains chronological records of event sequences, while the spatial memory module reconstructs environmental states. These complementary memory systems establish bidirectional connections with the planner-critic decision-making module through shared spatio-temporal belief representations. The Planner and Critic components collaboratively operate based on these unified beliefs: The Planner formulates subgoals and devises corresponding action plans, while the Critic simultaneously performs real-time validation of plan feasibility through dual criteria - environmental consistency check and temporal obsolescence detection. Upon identifying potential deficiencies, the Critic generates actionable refinements through structured feedback loops. This collaborative mechanism effectively addresses the inherent challenges associated with long-horizon planning tasks by maintaining dynamic alignment between memorialized environmental states, temporal context awareness, and situated action planning.

\section{Implementation Details}
\label{sec:appendix_alg}

\subsection{Main Algorithm}

The overall framework is outlined in Alg. \ref{alg:main_alg}.

\begin{algorithm}[t]
   \caption{STMA Execution Process}
   \label{alg:main_alg}
\begin{algorithmic}[1]
\REQUIRE Environment interface $\text{env}$, initial observation $o_1$, max steps $T$.
\ENSURE Task completion status.
\STATE Initialize environment: $\text{done} \leftarrow \text{False},\ i \leftarrow 1,\ f_1 \leftarrow \emptyset$; 
\STATE Initialize memory: $\mathcal{H} \leftarrow \emptyset,\ \mathcal{G} \leftarrow \emptyset$;
\WHILE{$i \leq T$ \AND $\neg \text{done}$}
   \STATE Record history: $\mathcal{H} \leftarrow \mathcal{H} \cup \{(f_i, o_i, a_i)\}$ \COMMENT{Update history buffer};
   \STATE Generate temporal belief: $b_i^t \leftarrow \mathcal{Sum}(\mathcal{H})$ \COMMENT{Temporal summarization};
   \STATE Extract spatial relations: $G' \leftarrow \mathcal{R}(b_i^t)$ \COMMENT{Relation retriever};
   \STATE Update knowledge graph: $\mathcal{G} \leftarrow \mathcal{G} \cup G'$ \COMMENT{Dynamic KG maintenance};
   \STATE Retrieve relevant subgraph: $\mathcal{G}_s \leftarrow \text{Retrieve}(\mathcal{G}, o_i)$ \COMMENT{Algorithm \ref{alg:retrieve algorithm}};
   \STATE Generate spatial belief: $b_i^s \leftarrow \mathcal{A}(\mathcal{G}_s)$ \COMMENT{Relation aggregation};
   \STATE Generate plan: $\{\hat{a}_{i:k}\}_{k=1}^m \leftarrow \text{Planner}(b_i^t, b_i^s, o_i)$;
   
   \FOR{$k = 1$ \textbf{to} $m$}
      \STATE Verify action: $(p_k, f_k) \leftarrow \text{Critic}(\hat{a}_{i:k}, b_i^t, b_i^s, o_i)$;
      \IF{$p_k = \text{False}$}
         \STATE Update feedback: $f_i \leftarrow f_k$;
         \STATE \textbf{break} \COMMENT{Re-planning triggered};
      \ENDIF
      \STATE Execute action: $a_i \leftarrow \hat{a}_{i:k}$;
      \STATE Observe environment: $(o_{i+1}, \text{done}, \text{info}) \leftarrow \text{env}(a_i)$;
      \STATE Update timestep: $i \leftarrow i + 1$;
      \IF{$\text{done}$}
         \STATE \textbf{break}
      \ENDIF
   \ENDFOR
\ENDWHILE
\STATE \textbf{return} $\text{done}$
\end{algorithmic}
\end{algorithm}

\subsection{K-hop}

After extracting the initial vertices, we utilize a K-hop graph search algorithm to perform relationship-based exploration. This involves retrieving all \(K\)-hop neighbors of the selected vertices within the KG. The algorithm enables semantically associated extraction, effectively performing ``associative reasoning" through semantic connections.

\begin{algorithm}[tb]
   \caption{K-Hop}
   \label{alg:k_hop}
\begin{algorithmic}
   \STATE {\bfseries Input:} Graph $\mathcal{G} = (V, E)$, vertex list $V_{\text{list}}$, integer $K$.
   \STATE Initialize $subgraph\_nodes \gets \emptyset$ \COMMENT{Set to store nodes in the subgraph};
   \FOR{each node in $V_{\text{list}}$}
      \IF{node is in $\mathcal{G}.V$}
         \STATE $neighbors \gets$ \texttt{single\_source\_shortest\_path\_length}($\mathcal{G}$, node, \texttt{cutoff}=$K$);
         \STATE $subgraph\_nodes \gets subgraph\_nodes \cup \texttt{keys}(neighbors)$ \COMMENT{Add neighbors within K hops to the subgraph};
      \ENDIF
   \ENDFOR
   \STATE {\bfseries Output:} $subgraph\_nodes$
\end{algorithmic}
\end{algorithm}

\clearpage  

\section{Sample Prompt}

\subsection{Summarizer}

\begin{lstlisting}
**Prompt:**

You are an intelligent agent tasked with analyzing a robot's interactions with its environment. The robot's activity is provided as a trajectory consisting of a series of **Observations** and **Actions**. Your goal is to summarize the robot's history up to the latest point, creating a detailed description of what the robot has done so far. This summary will be referred to as the robot's **"belief."**

**Trajectory Format:**

```
Observation:
[Details of the first observation]
Action:
[Details of the first action]

Observation:
[Details of the second observation]
Action:
[Details of the second action]

...

Observation:
[Details of the latest observation]
FeedBack:
[Details of the latest feedback]
```

*Note:* The last iteration includes a **FeedBack** section instead of an **Action** because it represents the most recent observation that the agent has not yet acted upon.

**Instructions:**

1. **Read the Trajectory:** Carefully review each **Observation** and the corresponding **Action** taken by the robot in the provided trajectory.

2. **Analyze Interactions:** For each pair of **Observation** and **Action**, understand the context and purpose of the robot's behavior. Consider how each action responds to the preceding observation.

3. **Exclude Latest Feedback:** Do not include the **FeedBack** section in your summary, as it represents the current state awaiting the robot's next action.

4. **Summarize the History:** Compile a comprehensive and detailed summary of the robot's actions and interactions based on the analyzed trajectory. Your summary should:
   - Describe the sequence of actions the robot has performed.
   - Highlight key behaviors, patterns, and any objectives the robot appears to be pursuing.
   - Provide insights into the robot's understanding and adaptation to its environment.

5. **Format the Summary:** Present the summary in clear, coherent English, ensuring that it accurately reflects the robot's historical behavior as derived from the trajectory.

6. Additional Knowledge for domain specific task:
   - Please First summarize each iteration for reference, then summarize them all. Plases Notice the objects and  entrance's direction of rooms.!
   - The cookbook in the kitchen is extremely important. Please record every detail of it. The agent need to follow the instruction from the cookbook!!!!
   - Please add a detailed cookbook record section
   - Please record all the direction movement action like "go xxx". This is essential
\end{lstlisting}

\subsection{Relation Retriever}

\begin{lstlisting}
You are an advanced AI agent tasked with extracting only the spatial relationships between objects in a passage. Your output should focus solely on spatial positions and directions without referencing actions, events, or other non-spatial information. The relationships between objects or rooms can involve various positional prepositions such as "on," "in," "at," "in front of," "behind," and relative directional terms. 

**Important:** Please track any changes in the position of items and only capture the latest spatial relationship. Use the sequence of actions to identify the most recent location of each item. For example, if an item like the "purple potato" is moved, only its final recorded location should be included in the output.

Respond with a list of RDF triples structured to include only the most recent spatial relationships between entities.

**Note:** **Do not include any comments or additional explanations in your output!!!** Provide only the list of RDF triples as specified.

Respond with a list of RDF triples structured to include only the most recent spatial relationships between entities.

Format:
Each triple should follow this format: `["object 1", "relation", "object 2"]`.
Use atomic objects for clarity.

Output the list in the following format, surrounded by triple backticks:

```
[
  ["object 1", "relation", "object 2"],
  ...
]
```

This will ensure a clear representation of only the latest spatial relationships based on the passage provided.

**One-shot sample:** 

input:
```
Paragraph:
I began the task with a clear directive: to find a recipe in the cookbook located in the kitchen and prepare a meal. My initial location was the bathroom, a room that, despite its familiarity, lacked any items of immediate use. The toilet was present but empty. I noted the exits and decided to move north.

Upon exiting the bathroom, I found myself in a corridor, a typical passageway with exits to the east, north, and south. I checked my inventory and confirmed that I was carrying a raw purple potato and a raw yellow potato. I chose to continue north.

The next room was a bedroom. It contained a bed, which was currently unoccupied. I noted the exits and proceeded east into the living room. This room, like the others, was standard and contained a sofa, which was also empty. I considered the exits and decided to go west, returning to the bedroom.

In the bedroom, I examined the bed and placed the purple potato on it. I then placed the yellow potato on the bed as well. With the potatoes stored, I moved east back into the living room and then south into the kitchen.

The kitchen was more promising. It contained a fridge, an oven, a massive table with a knife, a counter with a cookbook, and a stove. I examined the fridge, which was solid and closed, and then the table, which was unstable and held a knife. I opened the cookbook and read the recipe for a dish that required a purple potato. The instructions were straightforward: chop the purple potato, roast it, and prepare the meal.

With the recipe in mind, I returned to the living room and then went west to the bedroom. I retrieved the yellow potato from the bed and carried it back to the living room, placing it on the sofa. I then returned to the bedroom to retrieve the purple potato.

Back in the kitchen, I prepared to follow the recipe. I took the purple potato and used the knife from the table to chop it. Next, I placed the chopped potato on the stove and roasted it according to the cookbook's directions. Once the potato was roasted, I prepared the meal.

With the meal ready, I enjoyed the dish, fulfilling the initial directive to cook and eat a delicious meal.

Entity:
["bathroom", "corridor", "bedroom", "living room", "kitchen", "toilet", "purple potato", "yellow potato", "sofa", "fridge", "oven", "table", "knife", "counter", "cookbook", "stove"]
```

output:

```
[
    ["bathroom", "is at north of", "corridor"],
    ["corridor", "is at south of", "bathroom"],
    ["corridor", "is at north of", "bedroom"],
    ["bedroom", "is at south of", "corridor"],
    ["bedroom", "is at east of", "living room"],
    ["living room", "is at west of", "bedroom"],
    ["living room", "is at south of", "kitchen"],
    ["kitchen", "is at north of", "living room"],
    ["kitchen", "contains", "fridge"],
    ["kitchen", "contains", "oven"],
    ["kitchen", "contains", "table"],
    ["table", "holds", "knife"],
    ["kitchen", "contains", "counter"],
    ["counter", "holds", "cookbook"],
    ["kitchen", "contains", "stove"],
    ["sofa", "holds", "yellow potato"],
    ["stove", "holds", "purple potato"]
]
```
\end{lstlisting}

\subsection{Relation Aggregator}

\begin{lstlisting}
You are a smart agent with excellent spatial reasoning skills. I will provide you with a list of triples representing spatial relationships between objects. Your task is to reconstruct the environment and describe it in plain text.

**Input:** A list of interconnected tuples in the format `[obj1, relation, obj2]`.

**Output:** A coherent, factual description of the environment based on the given relationships.

Each tuple describes a key fact about the world, such as object locations or containment relationships. Your output should clearly explain the spatial layout and contents of the environment.

**Sample Input:**

```
[
    ["bathroom", "is at north of", "corridor"],
    ["corridor", "is at south of", "bathroom"],
    ["corridor", "is at north of", "bedroom"],
    ["bedroom", "is at south of", "corridor"],
    ["bedroom", "is at east of", "living room"],
    ["living room", "is at west of", "bedroom"],
    ["living room", "is at south of", "kitchen"],
    ["kitchen", "is at north of", "living room"],
    ["kitchen", "contains", "fridge"],
    ["kitchen", "contains", "oven"],
    ["kitchen", "contains", "table"],
    ["table", "holds", "knife"],
    ["kitchen", "contains", "counter"],
    ["counter", "holds", "cookbook"],
    ["kitchen", "contains", "stove"],
    ["sofa", "holds", "yellow potato"],
    ["stove", "holds", "purple potato"]
]
```

**Sample Output:**

The bathroom is located to the north of the corridor, which in turn is to the south of the bathroom. The corridor is also to the north of the bedroom, and the bedroom is to the south of the corridor. The bedroom is situated to the east of the living room, and the living room is to the west of the bedroom. The living room is located to the south of the kitchen, and the kitchen is to the north of the living room.

The kitchen contains several items: a fridge, an oven, a table, a counter, and a stove. The table holds a knife, and the counter holds a cookbook. Additionally, the sofa holds a yellow potato, and the stove holds a purple potato.

\end{lstlisting}

\subsection{Planner}
 
\begin{lstlisting}
**Role:**
You are an advanced AI operating as the central processor of a robot designed to interact with and adapt to real-world environments. Your main function is to assist in achieving user-defined goals by analyzing and interpreting environmental data. 

Please keep in mind the following key points:

1. **Dynamic Environment**: Each time you take an action, the environment will change. This means that the available actions and their meanings will shift after each step. It's crucial that you carefully consider how each current action will affect the environment, and how the available choices and their consequences will evolve after the action is executed. You are only provided with the current set of available actions-your understanding of what they mean should be grounded in the state of the environment at the time of decision-making.
2. **Plan Validity**: The plan you create is based on the current environment, but the environment is dynamic, and the set of available actions may change significantly during execution. This means your plan might quickly become outdated as you proceed. Always consider that actions may lead to changes in the environment that affect subsequent choices. Before finalizing a plan, ensure that you have thought through how each step might alter the available options and their implications.
3. **Exploration and Experimentation**: We encourage you to embrace exploration and learning through trial and error. Don't hesitate to test different approaches and learn from outcomes, as this will help refine your strategies and adapt to the unpredictable nature of real-world environments.
4. **Only generate action that is proper**: only valid action allowed, **No if statement action allowed!**

**Your task is:**
1. **Think through each step in a chain of thought** to determine the next useful subgoal that will help achieve the primary task.
2. **Based on what you have done and the spatial memory which describe the relationship of the object. Understand the overall environment (your belief).**  
3. **Propose a subgoal** based on the current environment and information, ensuring it contributes to the main goal.
4. **Plan a sequence of actions** that will complete the proposed subtask, drawing from the list of available actions.
5. **Ensure** that the sequence of actions effectively fulfills the subtask and progresses toward the overall goal.
6. **Fully Consider what spatial memory shows (This provide you with overall insight of the whole environment).**
7. **Get as much score as you can**: You can get 1 point if you:
    - 1. use correct action chop/cut/slice the correct ingredient follow the cookbook.
    - 2. use correct method fry/grill/roast to cook the ingredient follow the cookbook.
    - 3. collect correct ingredient to your inventory.

**Output Format:**
- **Subgoal:** Describe the specific subtask needed to advance the main goal.
- **Action Plan:** Select and arrange actions in a valid sequence that will complete this subtask.

**Thought:** You are suggested to think in this way (this is just a basic thought, you have to think in other way additionally):
1. What is your goal?
2. Base on the Spatio Memory, What situation you are facing?
3. Base on the What you have done, conclude an overall belief of you and envrionment.
4. What is your next subgoal that can achieve the task? Why?
    - Is this subgoal redundent?
    - What is the consequences when this subgoal completed?
    - Whether this subgoal fit **Every command** in your Knowledge.

5. Re-think whether it is a good subgoal. You can modify it. 
6. What is your plan to achieve the subgoal?
7. Please go through **each step** in your plan (each step is a selected_action), consider:
    - what may be the consequences of the action? 
    - Is this a proper action?
    - Is this a useless or redundent action?
    - Whether this action fit **Every command** in your Knowledge. 

8. As the environment change during your plan execution, please consider possible situation (including environment and possible actions) after each step executed.
... (more thought)

**Replan**: after go thought the thought step above, please think and review your previous plan, modify your plan base on the thought. This plan is your final plan.  

Use the following YAML format:

Thought:
1. ...
2. ...
3. ...
4. ...
5. ...
6. ...
...

Plan:
```YAML
Subgoal: "..." # The subgoal has to be detailed, including what you want to achieve, your strategies to achieve it. (and your plan action sequence in natural language)
Action Plan: 
    - "<selected_action 1>"
    - "<selected_action 2>"
    ...
    - "<selected_action n>"
```
\end{lstlisting}

\subsection{Critic}

\begin{lstlisting}
You are a highly advanced language model tasked with critically evaluating the suitability of a robot's proposed next action given its assigned subgoal, summarized execution history, and current environment. Your evaluation must carefully consider the following:
1. Dynamic Environment Awareness: Each action executed by the robot alters the environment. This means that the observations available, the significance of current candidate actions, and the consequences of executing an action will change with every step. When reasoning about the proposed action, reflect on its immediate significance in the current context and anticipate how the meaning and availability of future actions might evolve after execution. Your feedback should explicitly address this dynamic nature of the environment.
2. Encouragement of Exploration: While ensuring actions align with the subgoal and contribute meaningfully toward its completion, we also encourage the robot to explore and experiment. Exploration and trial-and-error are integral to discovering effective strategies, even if some actions may not yield immediate results. Highlight in your feedback how exploration may be beneficial in the current situation.
Your evaluation should provide a clear and reasoned analysis of the proposed action in the context of the subgoal and environment. Conclude with actionable feedback that integrates both the dynamic nature of the environment and the value of exploration.

You will be given the following information:

1. Robot's Subgoal: A clear description of the specific subgoal the robot is currently working toward.
2. Execution History: A summarized report detailing the robot's actions up to this point.
3. **spatial memory**: This memory record spatial relationship you have discoverd. Please Fully Consider what spatial memory shows (This provide you with overall insight of the whole environment).*
4. Knowledge: This is the most important part in your input. The Knowledge part will tell you how to achieve the goal!
5. Current Environment: The robot's current environment, showing details of its surroundings, relevant objects, and available paths.
6. All possible action: All possible action you can choose in this situation. Please judge the proposed action based on possible action choice.
7. Proposed Action: The action the robot is considering but has not yet executed.

Your responsibilities:

1. Reasoning:
   - Base on the Spatio Memory, What situation you are facing?
   - Base on the What you have done, conclude an overall belief of envrionment.
   - Analyze whether the subgoal is suitable according to the final task and current circumstances. (If not, you should told planner change  the subgoal)
   - Analyze the subgoal, execution history, current environment, and proposed action. 
   - Carefully assess whether the action is suitable and supports progress toward the subgoal. 
   - Consider the environmental context and prior actions to judge if the proposed action is logical and beneficial.
   - Consider whether this action is valid (in the possible action list) if not, please strictly reject the action.
   - Whether the movement fit **Every command** in your Knowledge.

2. Action Suitability: Explicitly indicate if the proposed action is suitable by selecting either `True` (action is relevant and beneficial) or `False` (action is unsuitable or potentially counterproductive).

3. Constructive Feedback: Provide a detailed explanation if the action is unsuitable, covering:
   - Why it was not suitable.
   - The key problem with the action in relation to the current environment and subgoal.
   - How to improve: Provide specific, actionable recommendations to adjust the robot's approach or action.
   - Any additional observations to help the robot advance toward the subgoal (or just suggest it to change the subgoal).
   - Please Encourage Exploration, Give less `False` prediction.

Output Format Requirements:
1. Internal Reasoning: Before providing the output, use a Chain of Thought approach to reason through your decision based on the subgoal, current state, and action suitability.
2. Structured Output: Follow this format:

```YAML
Action Suitability: True/False
Feedback: "Detailed explanation covering the reasons for suitability or unsuitability, key issues, actionable improvements, and additional insights for achieving the subgoal."
```
\end{lstlisting}

\end{document}